\documentclass{ecai} 
\usepackage{latexsym}
\usepackage{amssymb}
\usepackage{amsmath}
\usepackage{amsthm}
\usepackage{booktabs}
\usepackage{enumitem}
\usepackage{graphicx}
\usepackage{color}
\newtheorem{definition}{Definition}
\usepackage{booktabs,makecell, multirow, tabularx}
\usepackage{adjustbox}
\usepackage{bm}      % 加粗符号支持
\newcommand{\BibTeX}{B\kern-.05em{\sc i\kern-.025em b}\kern-.08em\TeX}

\begin{document}

%%%%%%%%%%%%%%%%%%%%%%%%%%%%%%%%%%%%%%%%%%%%%%%%%%%%%%%%%%%%%%%%%%%%%%%%

\begin{frontmatter}

%%% Use this command to specify your submission number.
%%% In doubleblind mode, it will be printed on the first page.

\paperid{7857} 

%%% Use this command to specify the title of your paper.

\title{A Modality-Tailored Graph Modeling Framework for Urban Region Representation via Contrastive Learning}
%%% Use this combinations of commands to specify all authors of your 
%%% paper. Use \fnms{} and \snm{} to indicate everyone's first names 
%%% and surname. This will help the publisher with indexing the 
%%% proceedings. Please use a reasonable approximation in case your 
%%% name does not neatly split into "first names" and "surname".
%%% Specifying your ORCID digital identifier is optional. 
%%% Use the \thanks{} command to indicate one or more corresponding 
%%% authors and their email address(es). If so desired, you can specify
%%% author contributions using the \footnote{} command.
\author[A]{Yaya Zhao}
\author[B]{Kaiqi Zhao}
\author[A]{Zixuan Tang} 
\author[A]{Zhiyuan Liu} 
\author[A]{Xiaoling Lu\thanks{Corresponding Author. Email: xiaolinglu@ruc.edu.cn.}}
\author[C]{Yalei Du} 

\address[A]{Center for Applied Statistics, School of Statistics, Innovation Platform, Renmin University of China}
\address[B]{Harbin Institute of Technology, Shenzhen}
\address[C]{Beijing Baixingkefu Network Technology Co., Ltd. \\ 
\texttt{\{zhaoyaya, 2021201741, 2023201970, xiaolinglu\}@ruc.edu.cn, zhaokaiqi@hit.edu.cn, yaleidu@163.com}}
% \address[A]{Center for Applied Statistics, School of Statistics, Innovation Platform, Renmin University of China}
% \address[B]{Harbin Institute of Technology, Shenzhen}
% \address[C]{Beijing Baixingkefu Network Technology Co., Ltd. \\ \email{\{zhaoyaya, 2021201741, 2023201970, xiaolinglu\}@ruc.edu.cn, zhaokaiqi@hit.edu.cn, yaleidu@163.com}}

%%% Use this environment to include an abstract of your paper.
\begin{abstract}
Graph-based models have emerged as a powerful paradigm for modeling multimodal urban data and learning region representations for various downstream tasks. However, existing approaches face two major limitations. (1) They typically employ identical graph neural network architectures across all modalities, failing to capture modality-specific structures and characteristics. (2) During the fusion stage, they often neglect spatial heterogeneity by assuming that the aggregation weights of different modalities remain invariant across regions, resulting in suboptimal representations. To address these issues, we propose MTGRR, a modality-tailored graph modeling framework for urban region representation, built upon a multimodal dataset comprising point of interest (POI), taxi mobility, land use, road element, remote sensing, and street view images. (1) MTGRR categorizes modalities into two groups based on spatial density and data characteristics: aggregated-level and point-level modalities. For aggregated-level modalities, MTGRR employs a mixture-of-experts (MoE) graph architecture, where each modality is processed by a dedicated expert GNN to capture distinct modality-specific characteristics. For the point-level modality, a dual-level GNN is constructed to extract fine-grained visual semantic features. (2) To obtain effective region representations under spatial heterogeneity, a spatially-aware multimodal fusion mechanism is designed to dynamically infer region-specific modality fusion weights. Building on this graph modeling framework, MTGRR further employs a joint contrastive learning strategy that integrates region aggregated-level, point-level, and fusion-level objectives to optimize region representations. Experiments on two real-world datasets across six modalities and three tasks demonstrate that MTGRR consistently outperforms state-of-the-art baselines, validating its effectiveness.
\end{abstract}
%%%%%%%%%%%%%%%%%%%%%%%%%%%%%%%%%%%%%%%%%%%%%%%%%%%%%%%%%%%%%%%%%%%%%%%%
\end{frontmatter}
\section{Introduction}
Graph-based models have demonstrated a strong ability to capture complex spatial dependencies and relational structures among urban regions. Recent works~\citep{Mandal_O’Connor_2024, AutoST, HREP} have leveraged this capability to learn comprehensive urban region representations from diverse data modalities, including points of interest (POIs), taxi trajectories, land use, road elements, remote sensing imagery, and street-view images. By integrating these multimodal data sources, the resulting region representations capture the distinct semantic and structural characteristics of each modality. These representations can be applied to various downstream tasks, including carbon emission estimation~\citep{Dai2023}, GDP prediction~\citep{Perera2024ImpactOE}, and population forecasting~\citep{Chen2022PopulationPO}.
\begin{figure}[tb]
    \centering
\includegraphics[width=0.87 \linewidth, height=6 cm]{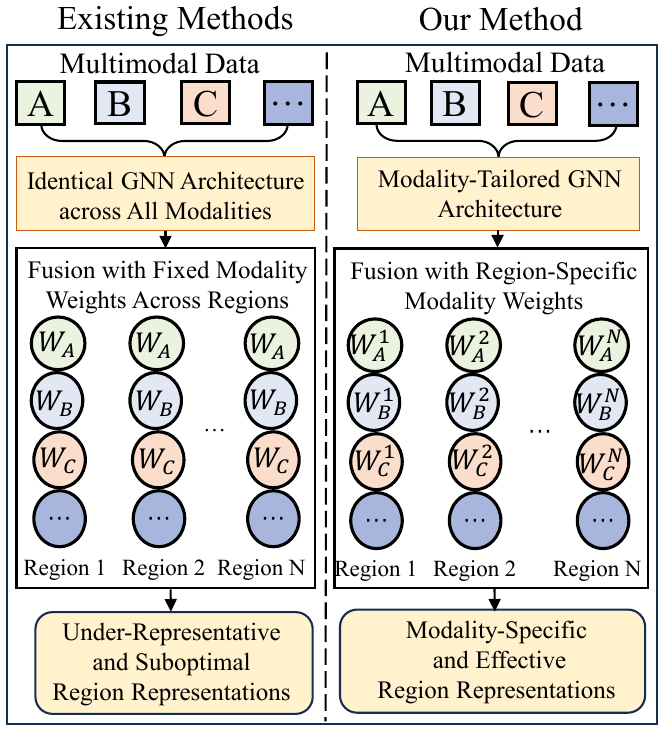}
    \caption{Comparison between existing methods and \textbf{MTGRR}.}
\label{fig:intro}
\end{figure}

Despite their effectiveness, existing graph-based models for urban region representation face two key limitations, as shown in Figure~\ref{fig:intro} (left). \textbf{Limitation 1:} Most approaches use identical graph neural network architectures across all modalities, applying uniform processing pipelines to heterogeneous data. 
For example, models such as \citep{CGAP,HREP} integrate POI distributions and taxi mobility patterns, which represent different aspects of urban dynamics, yet they still process them using the same GCN architecture \citep{Kipf_Welling}.
This modality-agnostic design leads to under-representative embeddings that fail to capture modality-specific structures and characteristics.  
\textbf{Limitation 2:} Existing modality fusion strategies often overlook spatial heterogeneity across regions, assuming that the contribution of each modality remains invariant across the city. For example,~\citep{AutoST, GraphST} directly average modality outputs to obtain region representations, and the fusion weights are shared across regions, failing to capture region-specific modality importance. This results in suboptimal representations that fail to reflect spatially varying urban characteristics.
% In practice, modality fusion weights are highly spatially dependent —for instance, street-view images tend to receive higher weights in central urban regions, whereas land use is often assigned greater importance in suburban regions.

To address these issues, we propose \textbf{MTGRR}, a modality-tailored graph framework for urban region representation (Figure~\ref{fig:intro}, right). It leverages a comprehensive urban dataset comprising six modalities, categorized into two groups based on data characteristics and spatial density.
\textbf{(1) Aggregated-level modalities} include POI, taxi mobility, land use, road element, and remote sensing imagery. Each POI is labeled with a functional category (e.g., restaurant, school); each taxi record is associated with a source region ID indicating inflow; each land parcel has a land-use label (e.g., residential, industrial); and each road segment is assigned a category label (e.g., living street, trunk link). For the first four modalities, modeling each point individually results in redundant information, as points with the same label exhibit highly similar features. Hence, they are better analyzed through label-based aggregation and treated as aggregated-level modalities. Although remote sensing imagery lacks explicit labels, it is considered aggregated-level because each region is associated with a single image that reflects region-level semantics.
\textbf{(2) Point-level modality} refers to street-view imagery, where each region contains numerous images capturing fine-grained visual semantics at specific spatial locations. Due to significant variation among individual images and the absence of well-defined predefined labels for grouping, these data are modeled at the instance level and treated as a point-level modality.

Based on this modality classification, \textbf{MTGRR} constructs modality-tailored graph neural networks for the two distinct types of urban modalities. For aggregated-level modalities, MTGRR adopts a mixture-of-experts (MoE) graph architecture, which consists of a global heterogeneous graph integrating all modalities, a set of expert GNNs each dedicated to a specific modality, and a modality-aware gating mechanism that adaptively fuses expert outputs into distinct modality-specific representations.  
For the point-level modality (i.e., street-view imagery), \textbf{MTGRR} employs a dual-level graph neural network to extract fine-grained visual semantic features. All images are treated as first-level nodes, and a second-level virtual node is introduced per region to aggregate local features. Intra-region edges connect all first-level image nodes to their corresponding second-level virtual node, while inter-region connections are only established between second-level nodes of adjacent regions. This hierarchical design enhances localized visual semantic representations.

Furthermore, to obtain effective region representations under spatial heterogeneity, \textbf{MTGRR} introduces a spatially-aware multimodal fusion mechanism that dynamically learns region-specific fusion weights, allowing the model to adaptively adjust modality contributions across diverse regions and enhance the fused representations. Building on this framework, MTGRR adopts a joint contrastive learning strategy that integrates aggregated-level, point-level, and fusion-level objectives
to optimize urban region representations.  
% Our main contributions are summarized as follows:
\begin{itemize}
\item We propose \textbf{MTGRR}, a modality-tailored graph modeling framework for urban region representation, which constructs a mixture-of-experts (MoE) graph architecture for aggregated-level modalities and a dual-level graph neural network for the point-level modality, enabling effective extraction of modality-specific representations incorporating distinct modality characteristics.
\item We introduce a spatially-aware multimodal fusion module that dynamically learns region-specific modality weights, enabling effective region representations that account for spatial heterogeneity.
\item We design a joint contrastive learning strategy that integrates region aggregated-level, point-level, and fusion-level contrastive objectives, jointly optimizing multimodal region representations.
\item We conduct extensive experiments on two real-world datasets covering six modalities and demonstrate that MTGRR consistently outperforms state-of-the-art methods across three downstream tasks, validating its effectiveness in diverse scenarios.
\end{itemize}
%%%%%%%%%%%%%%%%%%%%%%%%%%%%%%%%%%%%%%%%%%%%%%%%%%%%%%%%%%%%%%%%%%%%%%%%
\section{Related Work}
\paragraph{Graph Neural Networks for Multimodal Urban Data.}
Graph Neural Networks (GNNs) have shown strong potential in modeling multimodal urban data~\citep{schlichtkrull2018modeling}, 
% ~\citep{schlichtkrull2018modeling,wang2024relation,10.1007/978-3-031-70365-2_4}, 
where heterogeneous sources such as POIs, taxi mobility, remote sensing, and street-view imagery are naturally structured as graphs. Recent works leverage GNNs to jointly encode semantic and spatial information across modalities. 
Early approaches often relied on single-modality data, such as taxi mobility, to define urban graphs~\citep{wang2017region,MGFN,yao2018representing}, while more recent methods construct heterogeneous graphs from multiple modalities—e.g.,~\citep{luo2022urban,CGAP,ReMVC,HREP} combine POI and taxi mobility data to capture urban functionality and mobility patterns. Other studies~\citep{FlexiReg} further incorporate land use or remote sensing imagery to model physical infrastructure and environmental context. Meanwhile, contrastive learning has emerged as an effective tool for enhancing graph-based representations~\citep{Gong_Lin_Guo_Lin_Wang_Zheng_Zhou_Wan_2023,long2024clce,10.1007/978-3-031-70365-2_25}. Graph models~\citep{GURPP,AutoST} has been adopted to align multi-view embeddings through contrastive objectives.

\paragraph{Urban Region Representation Learning.}
Urban region representation learning has gradually evolved from single-modality modeling to more sophisticated multimodal fusion frameworks. Early approaches primarily focused on individual data sources such as taxi mobility~\citep{MGFN,yao2018representing}, POIs~\citep{huang2023learning}, or satellite imagery~\citep{urbanClip}, capturing only isolated aspects of urban dynamics. Recent studies increasingly incorporate multiple modalities to obtain richer and more comprehensive region representations~\citep{10.1145/3712698, li2023urban,Refound,zhang2021multi}. For example, Zhang et al.~\citep{zhang2021multi} propose a multi-view learning framework that combines taxi, POI, and check-in data to model regions from complementary perspectives.
Multimodal fusion methods vary considerably in design. Some adopt general attention mechanisms ~\citep{li2023urban,HAFusion,Refound}, while others leverage graph-based structures to encode spatial correlations and relational dependencies~\citep{GURPP,ReMVC,HREP}. 
In addition, recent models such as ~\citep{CGAL,MuseCL}  effectively integrate contrastive learning with multimodal fusion strategies to significantly enhance the quality and robustness of the learned region representations.
%%%%%%%%%%%%%%%%%%%%%%%%%%%%%%%%%%%%%%%%%%%%%%%%%%%%%%%%%%%%%%%%%%%%%%%%
\begin{figure*}[]
    \centering
\includegraphics[width=0.93\linewidth,height=3.41 in]{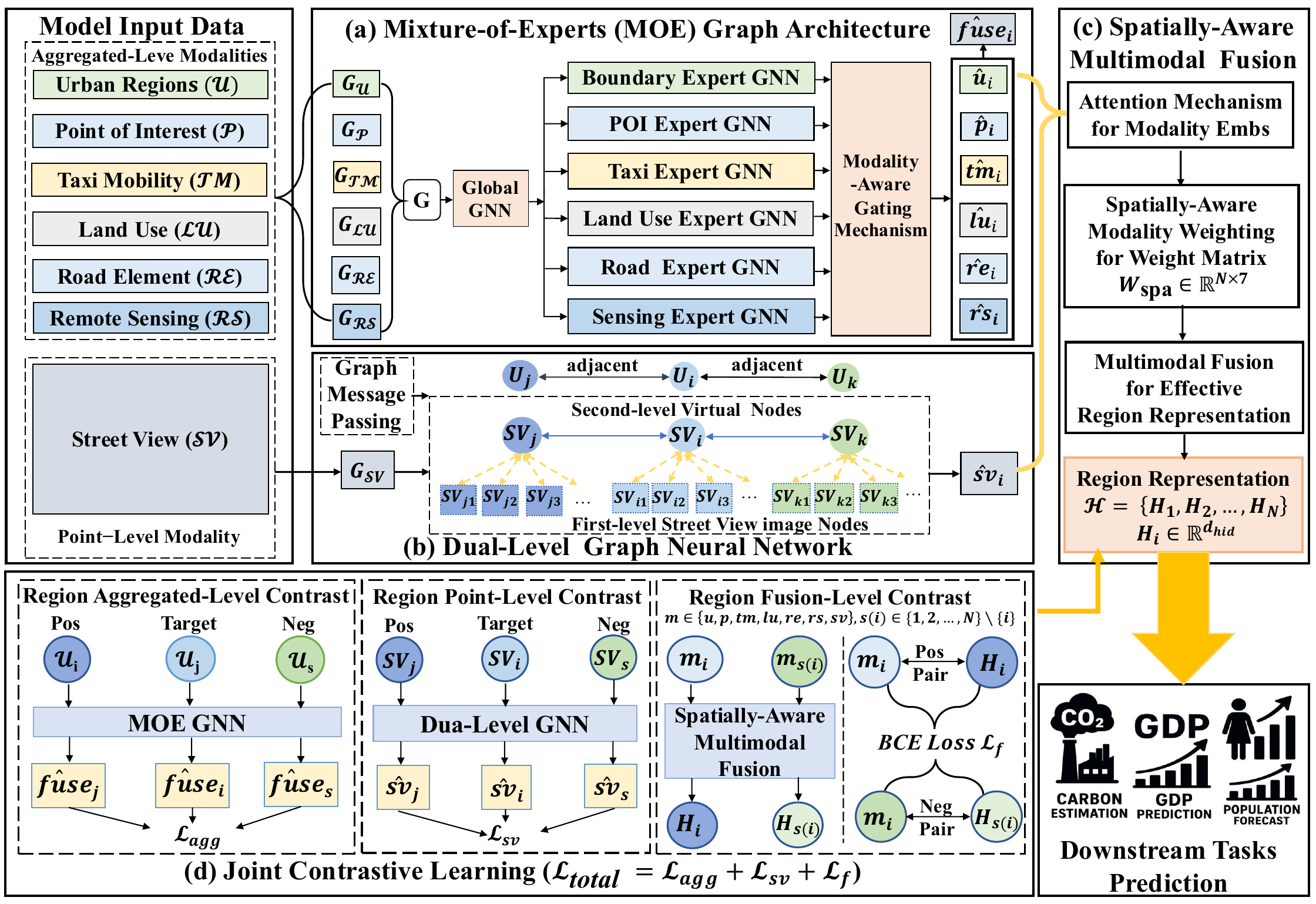}
\caption{
The framework: 
(a) MTGRR constructs a MoE graph architecture to model aggregated-level modalities, enabling distinct and accurate modality-specific representation. 
(b) A dual-level graph neural network is proposed to model the point-level street-view modality, capturing fine-grained visual semantic representations.
(c) A spatially-aware fusion mechanism is introduced to dynamically infer region-specific modality fusion weights and generate effective region representations. 
(d) Region-level, point-level, and fusion-level contrastive objectives are jointly optimized for enhancing region representations.
}
\label{fig:pipeline}
\end{figure*}

\section{Preliminaries \& Problem Statement}
\begin{definition}[Urban Regions ($\mathcal{U}$)]  
An urban area is partitioned into $N$ non-overlapping regions, denoted as $\mathcal{U} = \{{U}_1, {U}_2, \dots, {U}_N\}$. Each $U_i$ has a geographical boundary.
\end{definition}
\begin{definition}[Aggregated-Level Modalities]\label{definition:Aggregated-M}
A modality is defined as \emph{aggregated-level} if each data point has a semantic label, and data points sharing the same label exhibit high feature similarity, making point-wise modeling redundant and better suited for label-based aggregation; or only one data instance exists per region, naturally representing region-level semantics. 
\begin{itemize}
    \item \textbf{Point-of-Interest (POI) ($\mathcal{P}$):}  
    Each POI point is associated with a functional category label (e.g., restaurant, school). For each region $U_i$, the collection of POI data is denoted as $P_i$. Its feature vector is defined as $\text{P\_V}_i = \{P_{i1}, P_{i2}, \dots, P_{i|\mathcal{P}|}\}$, where $P_{ij}$ indicates the count of POIs in category $j$ for region $U_i$.
    
    \item \textbf{Taxi Mobility ($\mathcal{TM}$):}  
    Each taxi trip is associated with a source region label. For each region $U_i$, the collection of incoming taxi mobility flows is denoted as $TM_i$. Its feature vector is defined as $\text{TM\_V}_i = \{TM_{i1}, TM_{i2}, \dots, TM_{iN}\}$, where $TM_{ij}$ indicates the number of trips arriving from region $U_j$.

    \item \textbf{Land Use ($\mathcal{LU}$):}  
    Each land parcel is associated with a land use category label (e.g., residential, industrial). For each region $U_i$, the collection of land use data is denoted as $LU_i$. Its feature vector is defined as $\text{LU\_V}_i = \{LU_{i1}, LU_{i2}, \dots, LU_{i|\mathcal{LU}|}\}$, where $LU_{ij}$ indicates the count of land parcels in category $j$ for region $U_i$.

    \item \textbf{Road Element ($\mathcal{RE}$):}  
    Each road element has a road type label (e.g., trunk, motorway). For each region $U_i$, the collection of road element data is denoted as $RE_i$. Its feature vector is defined as $\text{RE\_V}_i = \{RE_{i1}, RE_{i2}, \dots, RE_{i|\mathcal{RE}|}\}$, where $RE_{ij}$ indicates the count of road elements in category $j$ for region $U_i$.

    \item \textbf{Remote Sensing Imagery ($\mathcal{RS}$):}  
    For each region $U_i$, a remote sensing image is available and denoted as $RS_i$. Its feature vector $\text{RS\_V}_i \in \mathbb{R}^{d_{{in}}}$ is extracted using an image encoder (e.g., EfficientNet-B4~\citep{tan2019efficientnet}).
\end{itemize}
\end{definition}

\begin{definition}[Point-Level Modality]\label{definition:Region-P}
A modality is defined as \emph{point-level} if each region contains many data points with fine-grained spatial semantics, high feature variability, and no suitable labels for aggregation. It is thus better modeled at the point level.
\begin{itemize}
    \item \textbf{Street View Imagery ($\mathcal{SV}$):}  
    Street-view images are abundant and unlabeled within each region, capturing fine-grained local appearances such as buildings and streetscapes. For each region $U_i$, a set of images $\{SV_{i1}, SV_{i2}, \dots, SV_{i|\mathcal{SV}_i|}\}$ is collected, where each $SV_{ij}$ denotes the $j$-th street view image within region $U_i$.
\end{itemize}
\end{definition}
\begin{definition}[Problem Statement: Urban Region Representation]\label{definition:URR}
Given urban regions $\mathcal{U} = \{ U_1, U_2, \dots, U_N \}$ associated with aggregated-level modalities $\{ \mathcal{P}, \mathcal{TM}, \mathcal{LU}, \mathcal{RE}, \mathcal{RS} \}$ and point-level modality $\mathcal{SV}$, the objective is to learn a low-dimensional embedding $H_i \in \mathbb{R}^{d_{{hid}}}$ for each region $U_i$, resulting in embeddings $\boldsymbol{\mathcal{H}} = \{ H_1, H_2, \dots, H_N \}$. 
These embeddings are then used to predict $K$ downstream tasks by applying a Ridge regression model, represented as $Y \in \mathbb{R}^{N \times K}$, which includes tasks such as carbon emission estimation, GDP prediction, and population forecasting.
\end{definition}
%%%%%%%%%%%%%%%%%%%%%%%%%%%%%%%%%%%%%%%%%%%%%%%%%%%%%%%%%%%%%%%%%%%%%%
\section{Methodology}  
We propose \textbf{MTGRR} (Figure~\ref{fig:pipeline}), a graph-based framework for urban region representation. 
(1) To address \textbf{Limitation 1}, MTGRR respectively designs tailored GNNs for aggregated-level and point-level modalities to capture modality-specific representations.  
(2) To address \textbf{Limitation 2}, it introduces a spatially-aware multimodal fusion mechanism to enable more effective region representations.
(3) Finally, MTGRR employs a joint contrastive learning strategy to further optimize and enhance region representations.
% In this section, we propose MTGRR, a graph framework for urban region representation. Its key components are illustrated in Figure~\ref{fig:pipeline}.
% (1) To address \textbf{Limitation 1} in Section 1, MTGRR respectively designs tailored graph neural network architectures for aggregated-level and point-level modalities to capture distinct and accurate modality-specific representations.  
% (2) To address \textbf{Limitation 2} in Section 1, a spatially-aware multimodal fusion mechanism is introduced to dynamically infer region-specific fusion weights, enabling more effective region representations.  
% Finally, based on the proposed graph modeling design, MTGRR adopts a joint contrastive learning strategy to optimize and enhance region representations.
\subsection{Mixture-of-Experts (MoE) Graph Architecture}\label{Subsection:MoME}
To learn distinct and accurate modality-specific representations for aggregated-level modalities, we design a MOE graph architecture. 
\subsubsection{Global Heterogeneous Graph Neural Network}\label{Subsubsection:heteGNN}
\paragraph{Heterogeneous Graph Construction $G = (\mathcal{V}, \mathcal{E})$.}
For each region $U_i$, according to Definition~\ref{definition:Aggregated-M}, it is associated with five region-aggregated modalities: a POI entity $P_i$, a taxi mobility entity $TM_i$, a land use entity $LU_i$, a road element entity $RE_i$, and a remote sensing entity $RS_i$. Each entity is represented by a feature vector, denoted as $P\_V_i$, $TM\_V_i$, $LU\_V_i$, $RE\_V_i$, and $RS\_V_i$, respectively. Based on this information, we first construct six subgraphs as follows:
\begin{itemize}
    \item \textbf{Region boundary graph} $G_{\mathcal{U}}$: nodes $\mathcal{V}_{\mathcal{U}} = \{U_1, U_2, \dots, U_N\}$, where each $U_i$ represents a region. An edge $(U_i, U_j) \in \mathcal{E}_{\mathcal{U}}$ is created if regions $U_i$ and $U_j$ are geographically adjacent.

    \item \textbf{POI graph} $G_{\mathcal{P}}$: nodes $\mathcal{V}_{\mathcal{P}} = \{P_1, P_2, \dots, P_N\}$. An edge $(P_i, P_j) \in \mathcal{E}_{\mathcal{P}}$ is created if $\text{cosine}(P\_V_i, P\_V_j) > \epsilon_{\mathcal{P}}$.

    \item \textbf{Taxi mobility graph} $G_{\mathcal{TM}}$: nodes $\mathcal{V}_{\mathcal{TM}} = \{TM_1, TM_2, \dots, TM_N\}$. An edge $(TM_i, TM_j) \in \mathcal{E}_{\mathcal{TM}}$ is created if $\text{cosine}(TM\_V_i, TM\_V_j) > \epsilon_{\mathcal{TM}}$.

    \item \textbf{Land use graph} $G_{\mathcal{LU}}$: nodes $\mathcal{V}_{\mathcal{LU}} = \{LU_1, LU_2, \dots, LU_N\}$. An edge $(LU_i, LU_j) \in \mathcal{E}_{\mathcal{LU}}$ is created if $\text{cosine}(LU\_V_i, LU\_V_j) > \epsilon_{\mathcal{LU}}$.

    \item \textbf{Road element graph} $G_{\mathcal{RE}}$: nodes $\mathcal{V}_{\mathcal{RE}} = \{RE_1, RE_2, \dots, RE_N\}$. An edge $(RE_i, RE_j) \in \mathcal{E}_{\mathcal{RE}}$ is created if $\text{cosine}(RE\_V_i, RE\_V_j) > \epsilon_{\mathcal{RE}}$.

    \item \textbf{Remote sensing graph} $G_{\mathcal{RS}}$: nodes $\mathcal{V}_{\mathcal{RS}} = \{RS_1, RS_2, \dots, RS_N\}$. An edge $(RS_i, RS_j) \in \mathcal{E}_{\mathcal{RS}}$ is created if $\text{cosine}(RS\_V_i, RS\_V_j) > \epsilon_{\mathcal{RS}}$.
\end{itemize}

Then, we build the final heterogeneous graph $G = (\mathcal{V}, \mathcal{E})$ by integrating all nodes and edges from the six subgraphs and introducing cross-modality connections. Specifically, for each region $U_i$ $(i = 1, 2, \dots, N)$, we add edges connecting $U_i$ to its modality-specific nodes $P_i$, $TM_i$, $LU_i$, $RE_i$, and $RS_i$ to enable information exchange across modalities.
\paragraph{Global Graph Neural Network.}
To capture cross-region and cross-modality interactions, we apply a global graph neural network over the heterogeneous graph $G = (\mathcal{V}, \mathcal{E})$. For each region boundary node $U_i$, POI node $P_i$, taxi mobility node $TM_i$, land use node $LU_i$, road element node $RE_i$, and remote sensing node $RS_i$, we randomly initialize their node representations as $\boldsymbol{u}_i$, $\boldsymbol{p}_i$, $\boldsymbol{tm}_i$, $\boldsymbol{lu}_i$, $\boldsymbol{re}_i$, and $\boldsymbol{rs}_i$, respectively, where all representations lie in $\mathbb{R}^{d_{{in}}}$. We use a multi-layer GAT model~\citep{velickovic2018graph} to update node embeddings through message passing. After $C$ layers, the updated embeddings are further refined via a feed-forward neural network $\text{FNN}(\cdot)$~\citep{vaswani2017attention}. The final outputs for all modality-specific nodes are denoted as $\{\overline{\boldsymbol{u}}_i, \overline{\boldsymbol{p}}_i, \overline{\boldsymbol{tm}}_i, \overline{\boldsymbol{lu}}_i, \overline{\boldsymbol{re}}_i, \overline{\boldsymbol{rs}}_i\}$, where each vector lies in $\mathbb{R}^{d_{{hid}}}$.

\subsubsection{Dedicated Expert Graph Neural Network}  
Then, to extract the unique semantic characteristics of each modality, we construct a separate modality-specific expert graph neural network for each of the six subgraphs: a region boundary expert GNN for $G_{\mathcal{U}}$, a POI expert GNN for $G_{\mathcal{P}}$, a taxi mobility expert GNN for $G_{\mathcal{TM}}$, a land use expert GNN for $G_{\mathcal{LU}}$, a road element expert GNN for $G_{\mathcal{RE}}$, and a remote sensing expert GNN for $G_{\mathcal{RS}}$.
Each expert GNN maintains its own parameters without sharing across modalities, allowing independent modeling of modality-specific structures. 

Taking the POI expert GNN for $G_{\mathcal{P}}$ as an example, the representation of node $P_i$ at the $l$-th layer is denoted as $\tilde{\boldsymbol{p}}_i^{(l)} \in \mathbb{R}^{d_{{hid}}}$. The initial node features are set as $\tilde{\boldsymbol{p}}_i^{(0)} = \overline{\boldsymbol{p}}_i$, where $\overline{\boldsymbol{p}}_i$ is the output of node $P_i$ from the global heterogeneous graph neural network. The edge representation for modality $\mathcal{P}$ at layer $l$, denoted as $e_{\mathcal{P}}^{(l)} \in \mathbb{R}^{d_{{hid}}}$, is randomly initialized. The update rules for the node and edge representations at the $l$-th layer of the POI expert GNN are as follows:
\begin{equation}\small
\begin{aligned}
\tilde{\boldsymbol{p}}_i^{(l)} &= \sigma\left( \sum_{j \in \mathcal{N}_{\mathcal{P}}(i) \cup \{i\}} \frac{\mathbf{W}_{\mathcal{P}}^{(l)}\left(\tilde{\boldsymbol{p}}_j^{(l-1)} \circ e_{\mathcal{P}}^{(l-1)}\right)}{\sqrt{|\mathcal{N}_{\mathcal{P}}(i)| \cdot |\mathcal{N}_{\mathcal{P}}(j)|}} \right), \\
e_{\mathcal{P}}^{(l)} &= \mathbf{E}_{\mathcal{P}}^{(l)} e_{\mathcal{P}}^{(l-1)} + \mathbf{b}_{\mathcal{P}}^{(l)},
\end{aligned}
\label{eq:poi_expert_update}
\end{equation}
where $\sigma(\cdot)$ denotes the activation function, $\mathbf{W}_{\mathcal{P}}^{(l)} \in \mathbb{R}^{d_{hid} \times d_{hid}}$, $\mathbf{E}_{\mathcal{P}}^{(l)} \in \mathbb{R}^{d_{hid} \times d_{hid}}$, and $\mathbf{b}_{\mathcal{P}}^{(l)} \in \mathbb{R}^{d_{hid}}$ are the learnable parameters for the POI expert GNN at the $l$-th layer, $\mathcal{N}_{\mathcal{P}}(i)$ denotes the index set of neighbors of node $P_i$ in $G_{\mathcal{P}}$, and $\circ$ denotes element-wise multiplication. After $L$ layers, the final output $\tilde{\boldsymbol{p}}_i^{(l)}$ for the POI node $P_i$ is represented as $\tilde{\boldsymbol{p}}_i \in \mathbb{R}^{d_{{hid}}}$. Similarly, the final outputs for each modality, computed using their corresponding expert GNNs, are denoted as $\tilde{\boldsymbol{u}}_i$, $\tilde{\boldsymbol{p}}_i$, $\tilde{\boldsymbol{tm}}_i$, $\tilde{\boldsymbol{lu}}_i$, $\tilde{\boldsymbol{re}}_i$, and $\tilde{\boldsymbol{rs}}_i$, respectively.

\subsubsection{Modality-Aware Gating Mechanism}
To dynamically fuse the expert outputs into distinct and accurate modality-specific representations, we design a modality-aware gating mechanism based on the feature vectors of each modality.

According to Definition~\ref{definition:Aggregated-M}, each region boundary node $U_i$ is associated with five aggregated-level modality nodes $\{P_i, TM_i, LU_i, RE_i, RS_i\}$, with corresponding feature vectors $\{P\_V_i, TM\_V_i, LU\_V_i, RE\_V_i, RS\_V_i\}$. In addition, the feature vector for $U_i$, denoted as $U\_V_i$, is initialized using Node2Vec embeddings~\citep{grover2016node2vec}. These feature vectors are separately projected into vectors in $\mathbb{R}^{d_{{hid}}}$ through modality-specific linear transformations.

The projected feature vectors are concatenated into a matrix $\text{{Concat}}\_V_i \in \mathbb{R}^{6 \times d_{{hid}}}$, which is passed through a gating network followed by a softmax function to produce the gating weights across the six modalities for region $U_i$, denoted as $\mathbf{g}_i \in \mathbb{R}^6$. These gating weights are used to modulate the outputs from the expert GNNs:
\begin{equation}\small
\hat{\boldsymbol{m}}_i = g_{i,m} \cdot \tilde{\boldsymbol{m}}_i, \quad \text{for each} \quad m \in \{u, p, tm, lu, re, rs\},
\end{equation}
where $g_{i,m}$ is the gating weight for modality $m$ in $\mathbf{g}_i$, and $\tilde{\boldsymbol{m}}_i$ is the corresponding expert GNN output. The gated output $\hat{\boldsymbol{m}}_i$ is obtained by scaling $\tilde{\boldsymbol{m}}_i$ with $g_{i,m}$, and is regarded as the final modality-specific representation produced by the MoE architecture.

\subsection{Dual-Level Graph Neural Network}\label{Subsection:DsvGNN}
In this subsection, to obtain fine-grained visual semantic representations for the point-level street-view modality, we design a dual-level graph neural network.
\subsubsection{Dual-Level Street-View Graph Construction}
According to Definition~\ref{definition:Region-P}, each region $U_i$ is associated with a set of street-view images $\{SV_{i1}, SV_{i2}, \dots, SV_{i|\mathcal{SV}_i|}\}$. To aggregate local street-view features, we introduce a virtual node $SV_i$ for each region $U_i$. Based on all image and virtual nodes, we construct a dual-level street-view graph $G_{\mathcal{SV}}$ for all regions as follows:

\begin{itemize}
    \item \textbf{Nodes of the street-view graph $G_{\mathcal{SV}}$:} 
    (1) \textbf{First-level nodes:} All street-view images $SV_{ij}$ $(i = 1, 2, \dots, N; \, j = 1, 2, \dots, |\mathcal{SV}_i|)$ are treated as individual \emph{first-level nodes}, where $i$ denotes region $U_i$ and $j$ indexes its associated images. 
    (2) \textbf{Second-level nodes:} All virtual nodes $SV_i$ $(i = 1, 2, \dots, N)$ are treated as \emph{second-level nodes}, capturing the aggregated region street-view features.
    
    \item \textbf{Edges of the street-view graph $G_{\mathcal{SV}}$:} 
    (1) \textbf{Intra-region edges:} For each region $U_i$ $(i = 1, 2, \dots, N)$, the virtual node $SV_i$ is connected to all its associated street-view image nodes $\{SV_{ij}\}$, where $j = 1, 2, \dots, |\mathcal{SV}_i|$. 
    (2) \textbf{Inter-region edges:} Two virtual nodes $SV_i$ and $SV_j$ are connected if their corresponding regions $U_i$ and $U_j$ are geographically adjacent.
\end{itemize}

\subsubsection{Graph Message Passing Mechanism}
To capture both intra- and inter-region interactions, we design a message passing mechanism for $G_{\mathcal{SV}}$. 
(1) Due to the large number of first-level image nodes, directly involving them in inter-region message passing would lead to an explosion of inter-region edges, significantly increasing the complexity of message propagation. Therefore, inter-region message passing occurs only between second-level virtual nodes. 
(2) Intra-region message passing occurs between each second-level virtual node and all first-level image nodes within the same region, enabling the capture of fine-grained visual semantics.

At the $l$-th layer, the representations of the first-level image nodes $SV_{ij}$ and the second-level virtual nodes $SV_i$ are denoted as $\hat{\boldsymbol{sv}}_{ij}^{(l)}$ and $\hat{\boldsymbol{sv}}_i^{(l)}$, respectively, each in $\mathbb{R}^{d_{{feat}}}$. The initial features $\hat{\boldsymbol{sv}}_{ij}^{(0)}$ are extracted using the CLIP-ViT-B/32 encoder~\citep{radford2021learning}. The initial feature of the second-level node $\hat{\boldsymbol{sv}}_i^{(0)}$ is obtained by averaging $\{\hat{\boldsymbol{sv}}_{ij}^{(0)}\}_{j=1}^{|\mathcal{SV}_i|}$.

Then at the $l$-th GNN layer, node features are updated as:
\begin{equation}\small
\begin{aligned}
\hat{\boldsymbol{sv}}_{ij}^{(l)} &= \sigma\left( \mathbf{W}_{sv1}^{(l)} \hat{\boldsymbol{sv}}_i^{(l-1)} \right),
\quad j = 1,2,\dots,|\mathcal{SV}_i|, \\
\hat{\boldsymbol{sv}}_i^{(l)} &= \sigma\left(
    \sum_{j=1}^{|\mathcal{SV}_i|} \mathbf{W}_{sv2}^{(l)} \hat{\boldsymbol{sv}}_{ij}^{(l-1)}
    + \sum_{s \in \mathcal{N}_{\mathcal{SV}}(i)} \mathbf{W}_{sv3}^{(l)} \hat{\boldsymbol{sv}}_s^{(l-1)}
\right),
\end{aligned}
\end{equation}
where $\sigma$ denotes the activation function. $\mathbf{W}_{sv1}^{(l)}$, $\mathbf{W}_{sv2}^{(l)}$, and $\mathbf{W}_{sv3}^{(l)} \in \mathbb{R}^{d_{{feat}} \times d_{{feat}}}$ are learnable.
After $Z$ layers, the final output 
$\hat{\boldsymbol{sv}}_i^{(l)}$ for the second-level virtual node $SV_i$ is passed through a feed-forward network ${\text{FNN}}(\cdot)$~\citep{vaswani2017attention} to obtain the final region-level street-view representation $\hat{\boldsymbol{sv}}_i \in \mathbb{R}^{d_{{hid}}}$ for each region $U_i$, where $i = 1, 2, \dots, N$.

\subsection{Spatially-Aware Multimodal Fusion Mechanism}\label{Subsection:SAFUSION}
To learn more effective region representations by considering spatial heterogeneity, we introduce a spatially-aware fusion mechanism that dynamically infers region-specific modality fusion weights. 
\paragraph{Attention Mechanism for Cross-Modal Interactions.}
For each region $U_i, i = 1, 2, \dots, N$, we use modality-specific embeddings $\{\hat{\boldsymbol{u}}_i, \hat{\boldsymbol{p}}_i, \hat{\boldsymbol{tm}}_i, \hat{\boldsymbol{lu}}_i, \hat{\boldsymbol{re}}_i, \hat{\boldsymbol{rs}}_i, \hat{\boldsymbol{sv}}_i\}$ obtained from the previous section. These embeddings are concatenated and fed into a multi-head attention module~\citep{vaswani2017attention} to model cross-modal interactions, producing unified modality embeddings $\boldsymbol{H}^{f} \in \mathbb{R}^{N \times 7 \times d_{{hid}}}$, where the second and third dimensions correspond to modality and hidden size.

\paragraph{Spatially-Aware Modality Weighting.}
To adaptively capture heterogeneity in modality contributions across diverse regions, we propose a dynamic spatially-aware weighting module that generates a region-specific modality weight matrix $\boldsymbol{W}_{\text{spa}} \in \mathbb{R}^{N \times 7}$, where each row corresponds to a region and each column to a modality.

Specifically, we first compute a region-level context vector $\boldsymbol{C} \in \mathbb{R}^{N \times d_{{hid}}}$ by averaging the cross-modal embeddings $\boldsymbol{H}^{f}$ across the modality dimension. This context vector is then passed through a lightweight gating network to generate attention scores as follows:
\begin{equation}\small
\boldsymbol{C} = \text{Average}(\boldsymbol{H}^{f}), \quad
\boldsymbol{O} = \sigma(\boldsymbol{C} \mathbf{W}_1) \mathbf{W}_2^\top, \quad
\boldsymbol{W}_{\text{spa}} = \text{Softmax}(\boldsymbol{O}),
\end{equation}
$\mathbf{W}_1 \in \mathbb{R}^{d_{{hid}} \times d_{{hid}}}$ and $\mathbf{W}_2 \in \mathbb{R}^{7 \times d_{{hid}}}$ are learnable parameters, $\sigma(\cdot)$ is a nonlinear activation (e.g., sigmoid), and softmax is applied over the modality dimension to normalize weights per region.

\paragraph{Multimodal Fusion for Effective Region Representation.}

Based on the spatial weights $\boldsymbol{W}_{\text{spa}}$, we compute the final region representations via the following multimodal fusion process:
\begin{equation}\small
\begin{array}{c}
\boldsymbol{H}^{w} = (\boldsymbol{W}_{\text{spa}} \odot \boldsymbol{H}^{f}) \mathbf{W}_{\text{proj}}, \quad
\boldsymbol{H}^{c} = p_1 \boldsymbol{H}^{w} + p_2 \boldsymbol{H}^{f}, \\
\boldsymbol{\mathcal{H}} = \text{Average}(\boldsymbol{H}^{c}, \text{dim}=1)
\end{array}
\end{equation}
where $\odot$ denotes element-wise multiplication along the modality dimension, $\mathbf{W}_{\text{proj}} \in \mathbb{R}^{d_{{hid}} \times d_{{hid}}}$ is learnable, and $p_1$, $p_2$ are learnable parameters. $\mathcal{H} = \{H_1, H_2, \dots, H_N\}$, with each $H_i \in \mathbb{R}^{d_{{hid}}}$, serves as the final region representation for downstream tasks.
%%%%%%%%%%%%%%%%%%%%%%%%%%%%%%%%%%%%%%%%%%%%%%%%%%%%%%%%%%%%%%%%%%%%%
\begin{table*}[!h]
\caption{Statistics of multimodal features and downstream tasks data across regions.}
\label{tab:dataset}
\centering
\setlength{\tabcolsep}{2 pt}
\resizebox{\textwidth}{!}{%
\begin{tabular}{c|c|c|c|c|c|c|c|c|c|c}
\toprule
\textbf{Dataset} & \textbf{Regions} & \textbf{POI} & \textbf{Taxi Mobility} & \textbf{Land Use} & \textbf{Road Element} & \textbf{Remote Sensing} & \textbf{Street View} & \textbf{Carbon} & \textbf{GDP} & \textbf{Pop} \\
\midrule
\textbf{Hangzhou} & 193 & 229682 & 418592 & 3984 & 16461 & 193 & 17072 & 385245 & 13021211 & 2151372 \\
\textbf{Shanghai} & 251 & 421000 & 106776 & 3916 & 22727 & 251 & 21755 & 844779 & 55343700 & 6898623 \\
\bottomrule
\end{tabular}}
\end{table*}
%%%%%%%%%%%%%%%%%%%%%%%%%%%%%%%%%%%%%%%%%%%%%%%%%%%%%%%%%%%%%%%%%%%%%
\begin{table*}[!h]
\caption{Performance comparison of baseline models across three tasks (carbon estimation, GDP prediction, and population forecasting) in two cities. 
$\mathcal{M}_1$--$\mathcal{M}_6$ represent different modality combinations, and \textbf{ALL} denotes using all modalities:
$\mathcal{M}_1 = (\mathcal{P}, \mathcal{TM})$,
$\mathcal{M}_2 = (\mathcal{RS}, \mathcal{SV})$,
$\mathcal{M}_3 = (\mathcal{P}, \mathcal{TM}, \mathcal{LU})$,
$\mathcal{M}_4 = (\mathcal{P}, \mathcal{TM}, \mathcal{RN}, \mathcal{RS})$,
$\mathcal{M}_5 = (\mathcal{P}, \mathcal{TM}, \mathcal{RS}, \mathcal{SV})$,
$\mathcal{M}_6 = (\mathcal{P}, \mathcal{LU}, \mathcal{RS}, \mathcal{SV})$,
\textbf{ALL} = $(\mathcal{P}, \mathcal{TM}, \mathcal{LU}, \mathcal{RN}, \mathcal{RS}, \mathcal{SV})$.
}
\label{tab:performance_comparison}
\centering
\setlength{\tabcolsep}{2 pt}
\resizebox{\textwidth}{!}{%
\begin{tabular}{l|c|ccc|ccc|ccc|ccc|ccc|ccc}
\toprule
\multirow{4}{*}{\textbf{Model}} & \multirow{4}{*}{$\mathcal{M}$} & \multicolumn{9}{c|}{\textbf{Hangzhou}} & \multicolumn{9}{c}{\textbf{Shanghai}} \\
\cmidrule(lr){3-11} \cmidrule(lr){12-20}
 & & \multicolumn{3}{c|}{\textbf{Carbon}} & \multicolumn{3}{c|}{\textbf{GDP}} & \multicolumn{3}{c|}{\textbf{Population}} & \multicolumn{3}{c|}{\textbf{Carbon}} & \multicolumn{3}{c|}{\textbf{GDP}} & \multicolumn{3}{c}{\textbf{Population}} \\
\cmidrule(lr){3-5} \cmidrule(lr){6-8} \cmidrule(lr){9-11} \cmidrule(lr){12-14} \cmidrule(lr){15-17} \cmidrule(lr){18-20}
 & & \textbf{MAE $\downarrow$} & \textbf{RMSE $\downarrow$} & \textbf{R2 $\uparrow$} & \textbf{MAE $\downarrow$} & \textbf{RMSE $\downarrow$} & \textbf{R2 $\uparrow$} & \textbf{MAE $\downarrow$} & \textbf{RMSE $\downarrow$} & \textbf{R2 $\uparrow$} & \textbf{MAE $\downarrow$} & \textbf{RMSE $\downarrow$} & \textbf{R2 $\uparrow$} & \textbf{MAE $\downarrow$} & \textbf{RMSE $\downarrow$} & \textbf{R2 $\uparrow$} & \textbf{MAE $\downarrow$} & \textbf{RMSE $\downarrow$} & \textbf{R2 $\uparrow$} \\
\midrule
\textbf{GraphST}\citep{GraphST} & $\mathcal{M}_1$ & 403.44 & 527.93 & 0.393 & 44619 & 62849 & 0.001 & 3338.2 & 3993.3 & 0.288 & 884.59 & 1167.36 & 0.291 & 119309 & 184082 & 0.251 & 6739.5 & 8241.9 & 0.365 \\
\textbf{ReCP}\citep{Recp} & $\mathcal{M}_1$ & 433.45 & 542.22 & 0.360 & 43789 & 59496 & 0.105 & 3276.5 & 3949.6 & 0.304 & 869.39 & 1074.09 & 0.400 & 115505 & 157699 & 0.450 & 5706.0 & 7207.2 & 0.515 \\
\textbf{HREP}\citep{HREP} & $\mathcal{M}_1$ & 340.72 & 434.12 & 0.590 & 37406 & 49575 & 0.378 & 2751.9 & 3445.6 & 0.470 & 629.97 & 849.28 & 0.625 & 95373 & 125970 & 0.649 & 4553.5 & 5945.6 & 0.670 \\
\textbf{UrbanVLP}\citep{UrbanVLP} & $\mathcal{M}_2$ & 428.79 & 544.21 & 0.355 & 39286 & 55508 & 0.221 & 3614.6 & 4384.5 & 0.142 & 733.48 & 943.20 & 0.537 & 109638 & 168654 & 0.371 & 5753.5 & 7325.3 & 0.498 \\
\textbf{HAFusion}\citep{HAFusion} & $\mathcal{M}_3$ & 406.06 & 525.56 & 0.398 & 38777 & 55164 & 0.230 & 3397.5 & 4167.9 & 0.225 & 699.75 & 917.77 & 0.562 & 89708 & 148511 & 0.512 & 5239.3 & 6739.2 & 0.576 \\
\textbf{GURPP}\citep{GURPP} & $\mathcal{M}_4$ & 299.84 & 383.90 & 0.679 & 27582 & 36957 & 0.655 & 2792.0 & 3446.1 & 0.470 & 720.08 & 949.27 & 0.531 & 96774 & 131216 & 0.619 & 6564.7 & 8131.1 & 0.382 \\
\textbf{MuseCL}\citep{MuseCL} & $\mathcal{M}_5$ & 440.22 & 568.45 & 0.296 & 50108 & 66063 & 0.003 & 3040.7 & 3756.7 & 0.370 & 938.63 & 1209.08 & 0.239 & 131563 & 192362 & 0.182 & 7106.1 & 8736.5 & 0.287 \\
\textbf{FlexiReg}\citep{FlexiReg} & $\mathcal{M}_6$ & 302.63 & 406.25 & 0.641 & 31887 & 44345 & 0.503 & 2453.9 & 3175.6 & 0.550 & 658.72 & 922.96 & 0.557 & 91837 & 133342 & 0.607 & 5054.1 & 6300.2 & 0.629 \\
\midrule
\multirow{7}{*}{\textbf{MTGRR}}  & $\mathcal{M}_1$ & 126.16 & 175.21 & 0.933 & 19583 & 28799 & 0.790 & 1922.2 & 2507.5 & 0.719 & 537.22 & 769.01 & 0.692 & 69455 & 97073 & 0.792 & 4112.5 & 5320.2 & 0.735 \\
 & $\mathcal{M}_2$ & \textbf{107.53} & \textbf{151.82} & \textbf{0.950} & 20950 & 28374 & 0.796 & 1813.2 & 2325.9 & 0.759 & 477.83 & 728.17 & 0.724 & 70275 & 97067 & 0.792 & 3957.4 & 5071.6 & 0.760 \\
 & $\mathcal{M}_3$ & 136.11 & 199.92 & 0.913 & 21126 & 29909 & 0.774 & 1899.8 & 2493.7 & 0.722 & 477.57 & 695.31 & 0.748 & 78269 & 103425 & 0.764 & 4191.6 & 5361.0 & 0.731 \\
 & $\mathcal{M}_4$ & 120.57 & 162.37 & 0.943 & 23737 & 34875 & 0.692 & 1851.4 & 2374.4 & 0.748 & 482.99 & 729.20 & 0.723 & 69317 & 98764 & 0.784 & 4258.6 & 5451.7 & 0.722 \\
 & $\mathcal{M}_5$ & 140.63 & 186.38 & 0.924 & 19607 & 27258 & 0.812 & {1688.8} & {2206.5} & {0.783} & 455.69 & 691.21 & 0.751 & 67890 & 91492 & 0.815 & 4093.8 & 5314.2 & 0.736 \\
 & $\mathcal{M}_6$ & 135.92 & 184.20 & 0.926 & 21901 & 30780 & 0.760 & 1835.2 & 2266.2 & 0.771 & 497.31 & 742.55 & 0.713 & 70974 & 94821 & 0.801 & 4016.1 & 5188.7 & 0.748 \\
 & ALL & {110.06} & 163.52 & {0.942} & \textbf{16994} & \textbf{23060} & \textbf{0.866} & \textbf{1681.8} & \textbf{2089.8} & \textbf{0.805} & \textbf{448.94} & \textbf{671.20} & \textbf{0.766} & \textbf{59782} & \textbf{84871} & \textbf{0.841} & \textbf{3851.1} & \textbf{4923.3} & \textbf{0.773} \\
\bottomrule
\end{tabular}}
\end{table*}
%%%%%%%%%%%%%%%%%%%%%%%%%%%%%%%%%%%%%%%%%%%%%%%%%%%%%%%%%%%%%%%%%%%%%
\begin{table*}[!h]
\caption{Performance evaluation of ablation experiments across three tasks (carbon estimation, GDP prediction, and population forecasting) in two cities.}
\label{tab:ablation_experiments}
\centering
\setlength{\tabcolsep}{2 pt}
\resizebox{\textwidth}{!}{%
\begin{tabular}{c|ccc|ccc|ccc|ccc|ccc|ccc}
\toprule
\multirow{3}{*}{\textbf{Method}} & \multicolumn{9}{c|}{\textbf{Hangzhou}} & \multicolumn{9}{c}{\textbf{Shanghai}} \\
\cmidrule(lr){2-10} \cmidrule(lr){11-19}
 & \multicolumn{3}{c|}{\textbf{Carbon}} & \multicolumn{3}{c|}{\textbf{GDP}} & \multicolumn{3}{c|}{\textbf{Population}} & \multicolumn{3}{c|}{\textbf{Carbon}} & \multicolumn{3}{c|}{\textbf{GDP}} & \multicolumn{3}{c}{\textbf{Population}} \\
\cmidrule(lr){2-4} \cmidrule(lr){5-7} \cmidrule(lr){8-10} \cmidrule(lr){11-13} \cmidrule(lr){14-16} \cmidrule(lr){17-19}
& \textbf{MAE $\downarrow$} & \textbf{RMSE $\downarrow$} & \textbf{R2 $\uparrow$} & \textbf{MAE $\downarrow$} & \textbf{RMSE $\downarrow$} & \textbf{R2 $\uparrow$} & \textbf{MAE $\downarrow$} & \textbf{RMSE $\downarrow$} & \textbf{R2 $\uparrow$} & \textbf{MAE $\downarrow$} & \textbf{RMSE $\downarrow$} & \textbf{R2 $\uparrow$} & \textbf{MAE $\downarrow$} & \textbf{RMSE $\downarrow$} & \textbf{R2 $\uparrow$} & \textbf{MAE $\downarrow$} & \textbf{RMSE $\downarrow$} & \textbf{R2 $\uparrow$} \\
\midrule
w/o MOE-GNN & 180.65 & 230.43 & 0.884 & 22185 & 30124 & 0.771 & 1978.3 & 2438.7 & 0.735 & 559.79 & 781.22 & 0.682 & 88511 & 118284 & 0.691 & 4740.4 & 6075.9 & 0.655 \\
w/o DL-GNN & 156.11 & 221.05 & 0.894 & 21517 & 28427 & 0.796 & 1927.8 & 2531.1 & 0.714 & 531.57 & 728.68 & 0.724 & 67107 & 88169 & 0.828 & 4469.9 & 5923.1 & 0.672 \\
w/o $\mathcal{SV}$ & 124.84 & 167.63 & 0.939 & 23298 & 34783 & 0.694 & 1996.7 & 2486.1 & 0.724 & 473.53 & 711.73 & 0.736 & 72195 & 95197 & 0.800 & 4293.6 & 5682.0 & 0.698 \\
w/o SA-MF & 155.43 & 209.15 & 0.905 & 25145 & 35275 & 0.685 & 1893.8 & 2398.6 & 0.743 & 457.21 & 740.19 & 0.715 & 73852 & 105153 & 0.756 & 4089.4 & 5188.2 & 0.748 \\
w/o $\mathcal{L}_{{sv}}$ & 129.46 & 179.81 & 0.930 & 22798 & 32941 & 0.726 & 2006.7 & 2616.4 & 0.694 & 478.96 & 694.16 & 0.749 & 70124 & 98117 & 0.787 & 4240.4 & 5307.9 & 0.737 \\
w/o $\mathcal{L}_{{f}}$ & 150.44 & 206.24 & 0.907 & 21118 & 29349 & 0.782 & 1841.1 & 2411.2 & 0.741 & 439.85 & 697.69 & 0.747 & 76225 & 104142 & 0.760 & 4067.0 & 5208.5 & 0.746 \\
MTGRR & \textbf{110.06} & \textbf{163.52} & \textbf{0.942} & \textbf{16994} & \textbf{23060} & \textbf{0.866} & \textbf{1681.8} & \textbf{2089.8} & \textbf{0.805} & \textbf{448.94} & \textbf{671.20} & \textbf{0.766} & \textbf{59782} & \textbf{84871} & \textbf{0.841} & \textbf{3851.1} & \textbf{4923.3} & \textbf{0.773} \\
\bottomrule
\end{tabular}
}
\end{table*}
%%%%%%%%%%%%%%%%%%%%%%%%%%%%%%%%%%%%%%%%%%%%%%%%%%%%%%%%%%%%%%%%%%%%%
\subsection{Joint Contrastive Learning}
To effectively optimize multimodal region representations, we introduce a joint contrastive learning strategy that integrates region aggregated-level, point-level, and fusion-level contrastive objectives.

\paragraph{Region Aggregated-Level Contrastive Learning.}
To enhance the effectiveness of aggregated-level modality representations, we introduce a region aggregated-level contrastive learning objective. 

Specifically, for each region $U_i$, we obtain embeddings $\{ \hat{\boldsymbol{u}}_i, \hat{\boldsymbol{p}}_i, \hat{\boldsymbol{tm}}_i, \hat{\boldsymbol{lu}}_i, \hat{\boldsymbol{re}}_i, \hat{\boldsymbol{rs}}_i \}$ for the aggregated-level modalities, each in $\mathbb{R}^{d_{{hid}}}$ in Subsection~\ref{Subsection:MoME}. These embeddings are concatenated and averaged to obtain a unified feature $\hat{\boldsymbol{fuse}}_i \in \mathbb{R}^{d_{{hid}}}$.
Given each anchor $\hat{\boldsymbol{fuse}}_i$, we select a positive sample $\hat{\boldsymbol{fuse}}_j$ from a geographically adjacent region $U_j$, and a negative sample $\hat{\boldsymbol{fuse}}_s$ from a non-adjacent region $U_s$. The contrastive loss is defined as:
\begin{equation}\small
\mathcal{L}_{{agg}} = \max\left( \|\hat{\boldsymbol{fuse}}_i - \hat{\boldsymbol{fuse}}_j\|_2 - \|\hat{\boldsymbol{fuse}}_i - \hat{\boldsymbol{fuse}}_s\|_2 + \gamma, 0 \right),
\end{equation}
where $\gamma$ is a margin hyperparameter and $\|\cdot\|_2$ denotes the L2 norm.

\paragraph{Region Point-Level Contrastive Learning.}
Similarly, to enhance the effectiveness of point-level street-view representations, we introduce a region point-level contrastive learning objective. 

Specifically, for each region $U_i$, we obtain a street-view embedding $\hat{\boldsymbol{sv}}_i \in \mathbb{R}^{d_{{hid}}}$ in Subsection~\ref{Subsection:DsvGNN}.
Given each anchor $\hat{\boldsymbol{sv}}_i$, a positive sample $\hat{\boldsymbol{sv}}_j$ is selected from a geographically adjacent region $U_j$, and a negative sample $\hat{\boldsymbol{sv}}_s$ from a non-adjacent region $U_s$. The region point-level contrastive loss is defined as:
\begin{equation}\small
\mathcal{L}_{{sv}} = \max\left( \|\hat{\boldsymbol{sv}}_i - \hat{\boldsymbol{sv}}_j\|_2 - \|\hat{\boldsymbol{sv}}_i - \hat{\boldsymbol{sv}}_s\|_2 + \gamma, 0 \right).
\end{equation}
\paragraph{Region Fusion-Level Contrastive Learning.}
To enhance the quality of fused region representations, we introduce a fusion-level contrastive objective based on binary cross-entropy (BCE) loss.

For each region $U_i$, we obtain its fused representation $\mathcal{H}_i \in \mathbb{R}^{d_{{hid}}}$ (Subsection~\ref{Subsection:SAFUSION}) and modality-specific embedding $\hat{\boldsymbol{m}}_i$ (Subsections~\ref{Subsection:MoME} or ~\ref{Subsection:DsvGNN}), where $i \in \{1, 2, \dots, N\}$ and $m \in \mathcal{A} = \{ \boldsymbol{u}, \boldsymbol{p}, \boldsymbol{tm}, \boldsymbol{lu}, \boldsymbol{re}, \boldsymbol{rs}, \boldsymbol{sv} \}$. For each $(i, m)$, a \emph{positive pair} (y = 1) is formed by matching $\hat{\boldsymbol{m}}_i$ with its corresponding $\mathcal{H}_i$, while a \emph{negative pair} (y = 0) is created by matching $\hat{\boldsymbol{m}}_i$ with a randomly sampled $\mathcal{H}_{s(i)}$ from a different region $U_{s(i)}$, where $s(i) \in \{1, 2, \dots, N\} \setminus \{i\}$. The fusion-level contrastive loss is defined as:
\begin{equation}\small
\mathcal{L}_{{f}} = - \sum_{i=1}^{N} \sum_{m \in \mathcal{A}} \left( 
\log \Phi([\hat{\boldsymbol{m}}_i, \mathcal{H}_i]) +
\log (1 - \Phi([\hat{\boldsymbol{m}}_i, \mathcal{H}_{s(i)}]))
\right),
\end{equation}
where $\Phi(\cdot)$ is a shared matching function implemented via a learnable MLP followed by a sigmoid activation.

\paragraph{Training.}
The final joint objective function is defined as:
\begin{equation}\small
\mathcal{L}_{{total}} = \mathcal{L}_{{agg}} + \mathcal{L}_{{sv}} + \mathcal{L}_{{f}}.
\end{equation}
This objective is used to optimize the entire modality-tailored graph modeling framework and obtain the final region representations.
\section{Evaluation}
\subsection{Experimental Setup}
\paragraph{Datasets and Metrics.} 
We use two real-world urban datasets from Hangzhou and Shanghai. The Shanghai dataset is based on \cite{MuseCL, urbanClip}, while the Hangzhou dataset is compiled by us. Each dataset covers a set of urban regions $\mathcal{U}$, with defined geographic boundaries. For each region, we collect six modalities: POI ($\mathcal{P}$), taxi mobility ($\mathcal{TM}$), land use ($\mathcal{LU}$), road element ($\mathcal{RN}$), remote sensing imagery ($\mathcal{RS}$), and street-view imagery ($\mathcal{SV}$). These datasets support three tasks: carbon emission estimation, GDP prediction, and population forecasting. Dataset statistics are summarized in Table~\ref{tab:dataset}. Performance is evaluated using Mean Absolute Error (MAE), Root Mean Squared Error (RMSE), and Coefficient of Determination ($R^2$).

\paragraph{Baselines.}
To evaluate the performance of MTGRR, we compare it with eight strong baselines:  
(1) \textbf{GraphST}~\citep{GraphST}, a self-supervised spatiotemporal graph model;  
(2) \textbf{ReCP}~\citep{Recp}, a framework for consistent multi-view representation learning;  
(3) \textbf{HREP}~\citep{HREP}, which incorporates heterogeneous region embedding with prompt learning;  
(4) \textbf{UrbanVLP}~\citep{UrbanVLP}, integrating satellite and street-view modalities via vision-language pretraining;  
(5) \textbf{HAFusion}~\citep{HAFusion}, using a dual-feature attentive fusion module to learn higher-order correlations;  
(6) \textbf{GURPP}~\citep{GURPP}, a graph-based region pretraining and prompting framework;  
(7) \textbf{MuseCL}~\citep{MuseCL}, a multi-semantic contrastive learning approach for region profiling;  
(8) \textbf{FlexiReg}~\citep{FlexiReg}, a flexible region encoder supporting adaptive formation and task-aware prompt learning.

\paragraph{Parameter Settings.}
All models are trained using the Adam optimizer (learning rate = 0.0001) for 300 epochs on both Hangzhou and Shanghai datasets. The contrastive margin $\gamma$ is fixed at 2 for both region aggregated-level and point-level objectives. For each modality-specific subgraph (see Subsection~\ref{Subsubsection:heteGNN}), the edge threshold $\epsilon$—including $\epsilon_{\mathcal{P}}, \epsilon_{\mathcal{TM}}, \epsilon_{\mathcal{LU}}, \epsilon_{\mathcal{RE}}, \epsilon_{\mathcal{RS}}$—is set as the 64th highest intra-modality cosine similarity. The node embedding dimension $d_{{feat}}$ of $G_{\mathcal{SV}}$ is set to 168, and the dual-level GNN uses one $Z{=}1$ layer. Additional hyperparameters are analyzed in Section~\ref{sec:param-analysis}. Experiments are conducted on a server with an Intel\textsuperscript{\textregistered} Xeon\textsuperscript{\textregistered} Gold 6148 CPU (80 cores, 2.40\,GHz) and a 24\,GB NVIDIA RTX 4090 GPU.
% 待添加参数设置内容
% ------------------------------------------------------------------------
\begin{figure*}[]
    \centering
    \includegraphics[width=1 \linewidth,height=1.2in]{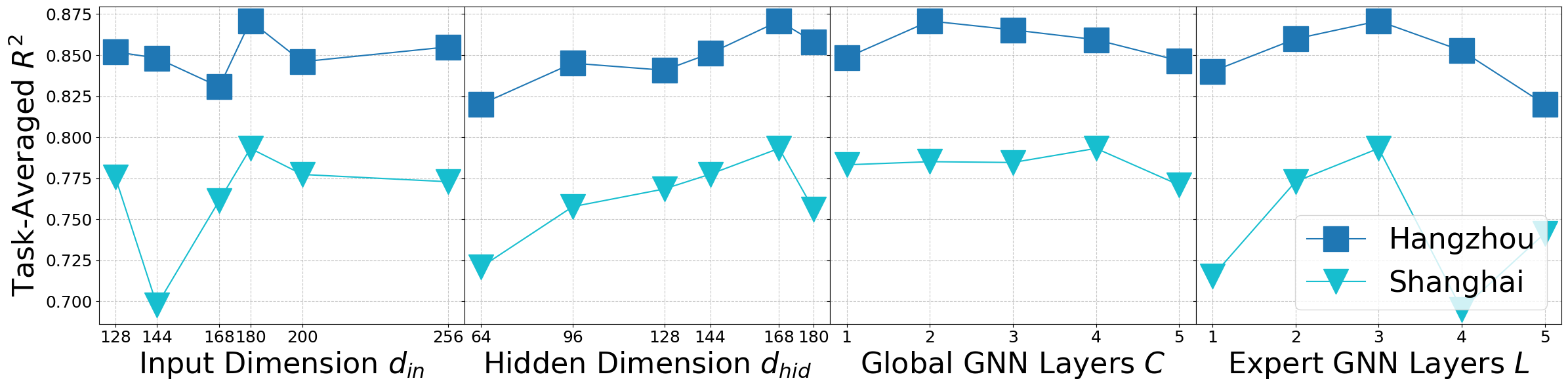}
    \caption{Hyperparameter study of MTGRR.} \label{fig:Effects of Hyperparameters}
\label{fig:hyperparam}
\end{figure*}
\begin{figure*}[]
    \centering
    \includegraphics[width=1 \linewidth,height=1.2in]{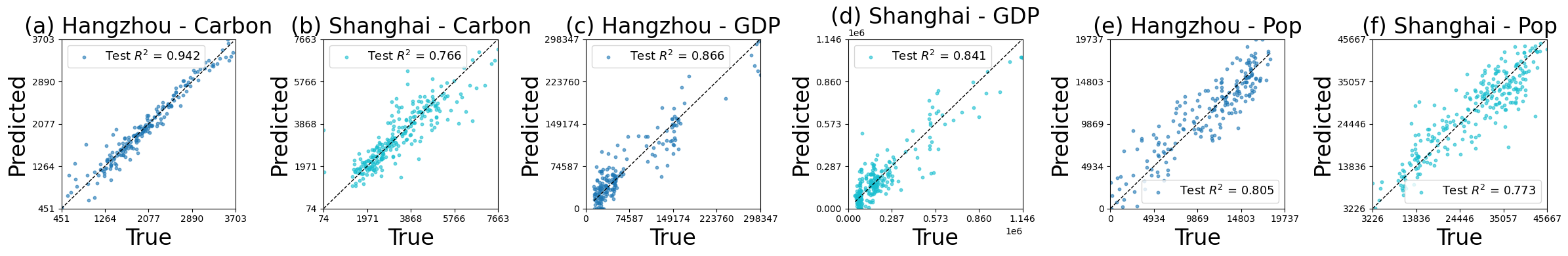}
   \caption{Predicted values generated by MTGRR vßersus ground truth on three downstream tasks across two cities.}
 \label{fig:Effects of Hyperparameters}
\label{fig:pred_vis}
\end{figure*}
 
\subsection{Overall Performance}
Table~\ref{tab:performance_comparison} summarizes the performance of MTGRR and eight strong baselines across three tasks in two cities. The baselines are evaluated under six modality combinations $\mathcal{M}_1$--$\mathcal{M}_6$, each representing a subset of the full modality set \textbf{ALL}, as shown in the column headers of Table~\ref{tab:performance_comparison}. We report MTGRR's performance under each subset and the full modality setting.
\textbf{(1) Modality-Specific Superiority:}
For each modality combination $\mathcal{M}_1$--$\mathcal{M}_6$, MTGRR consistently outperforms all baselines using the same subset across all tasks and cities, demonstrating the advantage of our modality-tailored graph modeling framework in learning modality-specific and effective representations.
\textbf{(2) Strong Performance with Limited Modalities:}
Even when using reduced modality combinations $\mathcal{M}_1$--$\mathcal{M}_6$, which are subsets of the full modality set \textbf{ALL}, MTGRR achieves strong results, often surpassing baselines that use more input data. This demonstrates its ability to maintain high accuracy with limited information, making it suitable for real-world scenarios where collecting comprehensive multimodal data is impractical or cost-prohibitive.
\textbf{(3) Scalability with More Modalities:}
MTGRR achieves its best performance under the full modality setting (\textbf{ALL}) in most cases, confirming that incorporating more modalities consistently and significantly improves predictive performance. The model demonstrates strong scalability and is able to effectively exploit diverse and complementary signals across modalities.
\textbf{(4) Robustness Across Modality Combinations:}
MTGRR maintains competitive performance across different modality combinations, tasks, and cities. This demonstrates its robustness in handling heterogeneous and incomplete modality inputs and its ability to flexibly adapt to diverse data distributions.

\subsection{Ablation Study}\label{ABstudy}
To assess the contribution of each component in MTGRR, we design six ablation variants:  
(1)~\textbf{w/o MOE-GNN}: Removes the expert GNNs and gating mechanism in the mixture-of-experts; each modality directly uses the output of the global GNN;  
(2)~\textbf{w/o DL-GNN}: Removes all first-level street-view nodes and their message passing; only uses the second-level node averaged from them, which is processed by a GAT for final region representation.
(3)~\textbf{w/o $\mathcal{SV}$}: Removes the street-view modality and all related modules;  
(4)~\textbf{w/o SA-MF}: Removes the spatially-aware multimodal fusion mechanism and applies multi-head attention over all modality embeddings without spatial weighting;  
(5)~\textbf{w/o $\mathcal{L}_{sv}$}: Removes the point-level contrastive loss;  
(6)~\textbf{w/o $\mathcal{L}_{f}$}: Removes the fusion-level contrastive loss.

Table~\ref{tab:ablation_experiments} shows that removing any individual component leads to performance degradation, confirming the importance of each module. Among all variants, \textbf{w/o MOE-GNN} and \textbf{w/o DL-GNN} result in the most significant drops, highlighting the critical role of our modality-tailored GNN designs for both aggregated and point-level modalities. Interestingly, \textbf{w/o DL-GNN} performs worse than \textbf{w/o $\mathcal{SV}$}, indicating that applying a standard GNN to street-view data is less effective than removing it entirely—underscoring the necessity of dual-level modeling for point-level visual signals. In addition, \textbf{w/o SA-MF} also leads to notable degradation, validating the importance of spatially-aware fusion in learning effective region representations. Lastly, the performance drops observed in \textbf{w/o $\mathcal{L}_{sv}$} and \textbf{w/o $\mathcal{L}_{f}$} confirm the benefits of joint contrastive learning.

\subsection{Hyperparameter Analysis}\label{sec:param-analysis}
We analyze the effect of four key hyperparameters in MTGRR: the input dimension $d_{{in}}$ for node representations in the global GNN, the hidden dimension $d_{{hid}}$ of modality representations used in multimodal fusion, the number of global GNN layers $C$, and the number of expert GNN layers $L$. As shown in Figure~\ref{fig:hyperparam}, we report the average $R^2$ across three prediction tasks—carbon emission, GDP, and population—in both Hangzhou and Shanghai. We observe that the best performance in both cities is consistently achieved with $d_{{in}} = 180$, $d_{{hid}} = 168$, and $L = 3$. The optimal global GNN depth is $C = 2$ for Hangzhou and $C = 4$ for Shanghai.

\subsection{Prediction Visualization and Fitting Ability}
To further assess the prediction quality of MTGRR, we visualize the predicted values versus the ground truth for three downstream tasks—carbon emission, GDP, and population—in both Hangzhou and Shanghai, as shown in Figure~\ref{fig:pred_vis}. Each subplot corresponds to a specific city-task pair and reports the $R^2$ score on the test set. We observe that the predicted values align closely with the ground truth, capturing both linear and nonlinear trends with high accuracy. The $R^2$ scores exceed 0.75 across all tasks and cities, demonstrating that the learned region representations are highly effective for prediction.

% \subsection{Efficiency Evaluation}
% Under identical hardware settings (see Sec. 5.1, Parameter Settings) and using the same Hangzhou dataset, we report training time (including data loading, preprocessing, and one-epoch training) for MTGRR and its eight baselines (see Sec. 5.1, Baselines)—with times (in seconds) of 95.14, 33.41, 70.26, 80.43, 153.74, 120.58, 81.82, 179.42, and 105.14, respectively. While some baselines \citep{GraphST,Recp,HREP,GURPP} train faster, their limited modality-specific modeling leads to lower accuracy. MTGRR offers a favorable trade-off, with slightly longer training time but significantly better performance. 

\section{Discussion and Conclusion}
We propose \textbf{MTGRR}, a modality-tailored graph framework for urban region representation. It integrates specialized GNNs and a spatially-aware fusion mechanism, jointly optimized via contrastive learning to derive effective region representations for downstream prediction tasks. Experiments across cities and tasks validate its superior performance and robustness. In future work, we will explore architectures beyond GNNs, such as large language models (LLMs), to better capture multimodal dependencies, and consider optimization methods beyond contrastive learning, including generative and reinforcement learning, to further enhance region representations.
%%%%%%%%%%%%%%%%%%%%%%%%%%%%%%%%%%%%%%%%%%%%%%%%%%%%%%%%%%%%%%%%%%%%%%
%%% Use this command to include your bibliography file.
\newpage
\section*{Acknowledgments}
This work is supported by the National Natural Science Foundation of China (No. 72171229), the MOE Project of the Key Research Institute of Humanities and Social Sciences (No. 22JJD110001), and the Big Data and Responsible Artificial Intelligence for National Governance at Renmin University of China. Yaya Zhao is supported by the "Qiushi Academic-Dongliang" Project of Renmin University of China (No. RUC24QSDL063).
\bibliography{mybibfile}

\begin{thebibliography}{36}
\providecommand{\natexlab}[1]{#1}
\providecommand{\url}[1]{\texttt{#1}}
\expandafter\ifx\csname urlstyle\endcsname\relax
  \providecommand{\doi}[1]{doi: #1}\else
  \providecommand{\doi}{doi: \begingroup \urlstyle{rm}\Url}\fi

\bibitem[Chen et~al.(2022)Chen, Mu, Li, and Dong]{Chen2022PopulationPO}
L.~Chen, T.~Mu, X.~Li, and J.~Dong.
\newblock Population prediction of chinese prefecture-level cities based on multiple models.
\newblock \emph{Sustainability}, 2022.
\newblock URL \url{https://api.semanticscholar.org/CorpusID:248303889}.

\bibitem[Chen et~al.(2025)Chen, Li, Jia, Shao, Zhao, Gao, Yang, and Yin]{10.1145/3712698}
M.~Chen, Z.~Li, H.~Jia, X.~Shao, J.~Zhao, Q.~Gao, M.~Yang, and Y.~Yin.
\newblock Mgrl4re: A multi-graph representation learning approach for urban region embedding.
\newblock \emph{ACM Trans. Intell. Syst. Technol.}, Jan. 2025.
\newblock ISSN 2157-6904.
\newblock \doi{10.1145/3712698}.
\newblock URL \url{https://doi.org/10.1145/3712698}.

\bibitem[Dai et~al.(2023)Dai, Zhou, Zhao, Li, and Liu]{Dai2023}
D.~Dai, B.~Zhou, S.~Zhao, K.~Li, and Y.~Liu.
\newblock Research on industrial carbon emission prediction and resistance analysis based on cei-egm-rm method: a case study of bengbu.
\newblock \emph{Scientific Reports}, 13\penalty0 (1):\penalty0 14528, 2023.
\newblock \doi{10.1038/s41598-023-41857-0}.
\newblock URL \url{https://doi.org/10.1038/s41598-023-41857-0}.

\bibitem[Gong et~al.(2023)Gong, Lin, Guo, Lin, Wang, Zheng, Zhou, and Wan]{Gong_Lin_Guo_Lin_Wang_Zheng_Zhou_Wan_2023}
L.~Gong, Y.~Lin, S.~Guo, Y.~Lin, T.~Wang, E.~Zheng, Z.~Zhou, and H.~Wan.
\newblock Contrastive pre-training with adversarial perturbations for check-in sequence representation learning.
\newblock \emph{Proceedings of the AAAI Conference on Artificial Intelligence}, 37\penalty0 (4):\penalty0 4276--4283, Jun. 2023.
\newblock \doi{10.1609/aaai.v37i4.25546}.
\newblock URL \url{https://ojs.aaai.org/index.php/AAAI/article/view/25546}.

\bibitem[Grover and Leskovec(2016)]{grover2016node2vec}
A.~Grover and J.~Leskovec.
\newblock node2vec: Scalable feature learning for networks.
\newblock In \emph{Proceedings of the 22nd ACM SIGKDD international conference on Knowledge discovery and data mining}, pages 855--864, 2016.

\bibitem[Hao et~al.(2025)Hao, Chen, Yan, Zhong, Wang, Wen, and Liang]{UrbanVLP}
X.~Hao, W.~Chen, Y.~Yan, S.~Zhong, K.~Wang, Q.~Wen, and Y.~Liang.
\newblock Urbanvlp: Multi-granularity vision-language pretraining for urban socioeconomic indicator prediction.
\newblock In \emph{Proceedings of the AAAI Conference on Artificial Intelligence}, volume~39, pages 28061--28069, 2025.
\newblock \doi{10.1609/aaai.v39i27.35024}.

\bibitem[Huang et~al.(2023)Huang, Zhang, Mai, Guo, and Cui]{huang2023learning}
W.~Huang, D.~Zhang, G.~Mai, X.~Guo, and L.~Cui.
\newblock Learning urban region representations with pois and hierarchical graph infomax.
\newblock \emph{ISPRS Journal of Photogrammetry and Remote Sensing}, 196:\penalty0 134--145, 2023.

\bibitem[Jin et~al.(2024)Jin, Song, Kan, Zhu, Sun, Li, Sun, and Zhang]{GURPP}
J.~Jin, Y.~Song, D.~Kan, H.~Zhu, X.~Sun, Z.~Li, X.~Sun, and J.~Zhang.
\newblock Urban region pre-training and prompting: A graph-based approach.
\newblock \emph{arXiv preprint arXiv:2408.05920}, 2024.

\bibitem[Kipf and Welling(2017)]{Kipf_Welling}
T.~N. Kipf and M.~Welling.
\newblock Semi-supervised classification with graph convolutional networks.
\newblock In \emph{International Conference on Learning Representations (ICLR)}, 2017.

\bibitem[Li et~al.(2023)Li, Huang, Cong, Wang, and Wang]{li2023urban}
Y.~Li, W.~Huang, G.~Cong, H.~Wang, and Z.~Wang.
\newblock Urban region representation learning with openstreetmap building footprints.
\newblock In \emph{Proceedings of the 29th ACM SIGKDD Conference on Knowledge Discovery and Data Mining}, pages 1363--1373, 2023.

\bibitem[Li et~al.(2024)Li, Huang, Zhao, Yang, Gong, and Chen]{Recp}
Z.~Li, W.~Huang, K.~Zhao, M.~Yang, Y.~Gong, and M.~Chen.
\newblock Urban region embedding via multi-view contrastive prediction.
\newblock In \emph{Proceedings of the AAAI Conference on Artificial Intelligence}, volume~38, pages 8724--8732, 2024.

\bibitem[Long et~al.(2024)Long, Zhuang, Killick, Meng, Mccreadie, and Aragon-Camarasa]{long2024clce}
Z.~Long, L.~Zhuang, G.~Killick, Z.~Meng, R.~Mccreadie, and G.~Aragon-Camarasa.
\newblock Clce: An approach to refining cross-entropy and contrastive learning for optimized learning fusion.
\newblock In \emph{ECAI 2024}, pages 1800--1807. IOS Press, 2024.

\bibitem[Luo et~al.(2022)Luo, Chung, and Chen]{luo2022urban}
Y.~Luo, F.-l. Chung, and K.~Chen.
\newblock Urban region profiling via multi-graph representation learning.
\newblock In \emph{Proceedings of the 31st ACM international conference on information \& knowledge management}, pages 4294--4298, 2022.

\bibitem[Mandal and O’Connor(2024)]{Mandal_O’Connor_2024}
S.~Mandal and N.~E. O’Connor.
\newblock Llmasmmkg: Llm assisted synthetic multi-modal knowledge graph creation for smart city cognitive digital twins.
\newblock \emph{Proceedings of the AAAI Symposium Series}, 4:\penalty0 210--221, 2024.

\bibitem[Perera and Fernando(2024)]{Perera2024ImpactOE}
P.~Perera and A.~Fernando.
\newblock Impact of energy prices and macroeconomic variables on gdp prediction uk: Machine learning approach.
\newblock \emph{Journal of Business and Management Studies}, 2024.
\newblock URL \url{https://api.semanticscholar.org/CorpusID:273115300}.

\bibitem[Radford et~al.(2021)Radford, Kim, Hallacy, Ramesh, Goh, Agarwal, Sastry, Askell, Mishkin, Clark, et~al.]{radford2021learning}
A.~Radford, J.~W. Kim, C.~Hallacy, A.~Ramesh, G.~Goh, S.~Agarwal, G.~Sastry, A.~Askell, P.~Mishkin, J.~Clark, et~al.
\newblock Learning transferable visual models from natural language supervision.
\newblock In \emph{International conference on machine learning}, pages 8748--8763. PMLR, 2021.

\bibitem[Schlichtkrull et~al.(2018)Schlichtkrull, Kipf, Bloem, Van Den~Berg, Titov, and Welling]{schlichtkrull2018modeling}
M.~Schlichtkrull, T.~N. Kipf, P.~Bloem, R.~Van Den~Berg, I.~Titov, and M.~Welling.
\newblock Modeling relational data with graph convolutional networks.
\newblock In \emph{The semantic web: 15th international conference, ESWC 2018}, pages 593--607. Springer, 2018.

\bibitem[Shui et~al.(2024)Shui, Li, Qi, Jiang, and Yu]{10.1007/978-3-031-70365-2_25}
C.~Shui, X.~Li, J.~Qi, G.~Jiang, and Y.~Yu.
\newblock Hierarchical graph contrastive learning for review-enhanced recommendation.
\newblock In A.~Bifet, J.~Davis, T.~Krilavi{\v{c}}ius, M.~Kull, E.~Ntoutsi, and I.~{\v{Z}}liobait{\.{e}}, editors, \emph{Machine Learning and Knowledge Discovery in Databases. Research Track}, pages 423--440, Cham, 2024. Springer Nature Switzerland.
\newblock ISBN 978-3-031-70365-2.

\bibitem[Sun et~al.(2024)Sun, Qi, Chang, Fan, Karunasekera, and Tanin]{HAFusion}
F.~Sun, J.~Qi, Y.~Chang, X.~Fan, S.~Karunasekera, and E.~Tanin.
\newblock Urban region representation learning with attentive fusion.
\newblock In \emph{2024 IEEE 40th International Conference on Data Engineering (ICDE)}, pages 4409--4421. IEEE, 2024.

\bibitem[Sun et~al.(2025)Sun, Chang, Tanin, Karunasekera, and Qi]{FlexiReg}
F.~Sun, Y.~Chang, E.~Tanin, S.~Karunasekera, and J.~Qi.
\newblock Urban region representation learning: A flexible approach, 2025.
\newblock URL \url{https://arxiv.org/abs/2503.09128}.

\bibitem[Tan and Le(2019)]{tan2019efficientnet}
M.~Tan and Q.~Le.
\newblock Efficientnet: Rethinking model scaling for convolutional neural networks.
\newblock In \emph{International conference on machine learning}, pages 6105--6114. PMLR, 2019.

\bibitem[Vaswani et~al.(2017)Vaswani, Shazeer, Parmar, Uszkoreit, Jones, Gomez, Kaiser, and Polosukhin]{vaswani2017attention}
A.~Vaswani, N.~Shazeer, N.~Parmar, J.~Uszkoreit, L.~Jones, A.~N. Gomez, {\L}.~Kaiser, and I.~Polosukhin.
\newblock Attention is all you need.
\newblock In \emph{Advances in Neural Information Processing Systems (NeurIPS)}, pages 5998--6008, 2017.

\bibitem[Velickovic et~al.(2018)Velickovic, Cucurull, Casanova, Romero, Li{\`o}, and Bengio]{velickovic2018graph}
P.~Velickovic, G.~Cucurull, A.~Casanova, A.~Romero, P.~Li{\`o}, and Y.~Bengio.
\newblock Graph attention networks.
\newblock In \emph{Proceedings of the 6th International Conference on Learning Representations (ICLR)}, 2018.

\bibitem[Wang and Li(2017)]{wang2017region}
H.~Wang and Z.~Li.
\newblock Region representation learning via mobility flow.
\newblock In \emph{Proceedings of the 2017 ACM on Conference on Information and Knowledge Management}, pages 237--246, 2017.

\bibitem[Wu et~al.(2022)Wu, Yan, Fan, Pan, Zhu, Zheng, Cheng, and Wang]{MGFN}
S.~Wu, X.~Yan, X.~Fan, S.~Pan, S.~Zhu, C.~Zheng, M.~Cheng, and C.~Wang.
\newblock Multi-graph fusion networks for urban region embedding.
\newblock In \emph{Proceedings of the Thirty-First International Joint Conference on Artificial Intelligence, {IJCAI-22}}, pages 2312--2318, 2022.

\bibitem[Xiao et~al.(2024)Xiao, Zhou, Xiao, Huang, and Xiong]{Refound}
C.~Xiao, J.~Zhou, Y.~Xiao, J.~Huang, and H.~Xiong.
\newblock Refound: Crafting a foundation model for urban region understanding upon language and visual foundations.
\newblock In \emph{Proceedings of the 30th ACM SIGKDD Conference on Knowledge Discovery and Data Mining}, 2024.
\newblock ISBN 9798400704901.

\bibitem[Xu and Zhou(2024)]{CGAP}
Z.~Xu and X.~Zhou.
\newblock Cgap: Urban region representation learning with coarsened graph attention pooling.
\newblock In K.~Larson, editor, \emph{Proceedings of the Thirty-Third International Joint Conference on Artificial Intelligence, {IJCAI-24}}, pages 7518--7526, 2024.

\bibitem[Yan et~al.(2024)Yan, Wen, Zhong, Chen, Chen, Wen, Zimmermann, and Liang]{urbanClip}
Y.~Yan, H.~Wen, S.~Zhong, W.~Chen, H.~Chen, Q.~Wen, R.~Zimmermann, and Y.~Liang.
\newblock Urbanclip: Learning text-enhanced urban region profiling with contrastive language-image pretraining from the web.
\newblock In \emph{Proceedings of the ACM Web Conference 2024}, WWW '24, page 4006–4017, 2024.

\bibitem[Yao et~al.(2018)Yao, Fu, Liu, Hu, and Xiong]{yao2018representing}
Z.~Yao, Y.~Fu, B.~Liu, W.~Hu, and H.~Xiong.
\newblock Representing urban functions through zone embedding with human mobility patterns.
\newblock In \emph{Proceedings of the Twenty-Seventh International Joint Conference on Artificial Intelligence (IJCAI-18)}, 2018.

\bibitem[Yong and Zhou(2024)]{MuseCL}
X.~Yong and X.~Zhou.
\newblock Musecl: Predicting urban socioeconomic indicators via multi-semantic contrastive learning.
\newblock In K.~Larson, editor, \emph{Proceedings of the Thirty-Third International Joint Conference on Artificial Intelligence, {IJCAI-24}}, pages 7536--7544, 2024.

\bibitem[Zhang et~al.(2022)Zhang, Long, and Cong]{ReMVC}
L.~Zhang, C.~Long, and G.~Cong.
\newblock Region embedding with intra and inter-view contrastive learning.
\newblock \emph{IEEE Transactions on Knowledge and Data Engineering}, 35\penalty0 (9):\penalty0 9031--9036, 2022.

\bibitem[Zhang et~al.(2021)Zhang, Li, Li, and Hui]{zhang2021multi}
M.~Zhang, T.~Li, Y.~Li, and P.~Hui.
\newblock Multi-view joint graph representation learning for urban region embedding.
\newblock In \emph{Proceedings of the twenty-ninth international conference on international joint conferences on artificial intelligence}, pages 4431--4437, 2021.

\bibitem[Zhang et~al.(2023{\natexlab{a}})Zhang, Huang, Xia, Wang, Li, and Yiu]{AutoST}
Q.~Zhang, C.~Huang, L.~Xia, Z.~Wang, Z.~Li, and S.~Yiu.
\newblock Automated spatio-temporal graph contrastive learning.
\newblock In \emph{Proceedings of the ACM Web Conference 2023}, pages 295--305, 2023{\natexlab{a}}.

\bibitem[Zhang et~al.(2023{\natexlab{b}})Zhang, Huang, Xia, Wang, Yiu, and Han]{GraphST}
Q.~Zhang, C.~Huang, L.~Xia, Z.~Wang, S.~M. Yiu, and R.~Han.
\newblock Spatial-temporal graph learning with adversarial contrastive adaptation.
\newblock In \emph{International Conference on Machine Learning}, pages 41151--41163. PMLR, 2023{\natexlab{b}}.

\bibitem[Zhang et~al.(2019)Zhang, Fu, Wang, Li, and Zheng]{CGAL}
Y.~Zhang, Y.~Fu, P.~Wang, X.~Li, and Y.~Zheng.
\newblock Unifying inter-region autocorrelation and intra-region structures for spatial embedding via collective adversarial learning.
\newblock In \emph{Proceedings of the 25th ACM SIGKDD International Conference on Knowledge Discovery \& Data Mining}, pages 1700--1708, 2019.

\bibitem[Zhou et~al.(2023)Zhou, He, Chen, Shang, and Han]{HREP}
S.~Zhou, D.~He, L.~Chen, S.~Shang, and P.~Han.
\newblock Heterogeneous region embedding with prompt learning.
\newblock In \emph{Proceedings of the AAAI Conference on Artificial Intelligence}, volume~37, pages 4981--4989, 2023.

\end{thebibliography}
\end{document}